\pdfoutput=1
\documentclass[conference]{IEEEtran}
\IEEEoverridecommandlockouts
\usepackage{cite}
\usepackage{amsmath,amssymb,amsfonts}
\usepackage{algorithmic}
\usepackage{textcomp}
\usepackage{times}
\usepackage{epsfig}
\usepackage{graphicx}
\usepackage{amsmath}
\usepackage{amssymb}

\usepackage{multirow}
\usepackage{tabularx}
\usepackage{array}
\usepackage{tabu}
\usepackage{adjustbox, amsmath}

\usepackage{makecell}

\usepackage{pifont}

\usepackage[table, xcdraw]{xcolor}
\usepackage{booktabs}
\usepackage{hhline}

\newcommand{\Ds}{\mathcal{D}_{S}}
\newcommand{\Dt}{\mathcal{D}_{T}}

\newcommand{\etal}{\textit{et al}.}

\def\BibTeX{{\rm B\kern-.05em{\sc i\kern-.025em b}\kern-.08em
    T\kern-.1667em\lower.7ex\hbox{E}\kern-.125emX}}
\begin{document}

\makeatletter
\newcommand{\linebreakand}{%
  \end{@IEEEauthorhalign}
  \hfill\mbox{}\par
  \mbox{}\hfill\begin{@IEEEauthorhalign}
}
\makeatother

\title{Few-Shot Object Detection in Unseen Domains
\thanks{$^*$Both authors have contributed equally to this work.}\\
\thanks{$^\dagger$ Author is also affiliated with Karlsruhe Institute of Technology (KIT)}
}

\author{
\IEEEauthorblockN{Karim Guirguis$^{* \dagger}$}
\IEEEauthorblockA{\textit{Robert Bosch Corporate Research} \\
Renningen, Germany \\
karim.guirguis@de.bosch.com}
\and
\IEEEauthorblockN{George Eskandar$^*$}
\IEEEauthorblockA{\textit{University Of Stuttgart} \\
Stuttgart, Germany \\
george.eskandar@iss.uni-stuttgart.de}
\and
\IEEEauthorblockN{Matthias Kayser}
\IEEEauthorblockA{\textit{Robert Bosch Corporate Research} \\
Renningen, Germany \\
matthias.ochs2@de.bosch.com}
\linebreakand 
\IEEEauthorblockN{Bin Yang}
\IEEEauthorblockA{\textit{University Of Stuttgart} \\
Stuttgart, Germany \\
bin.yang@iss.uni-stuttgart.de}
\and
\IEEEauthorblockN{Juergen Beyerer$^\dagger$}
\IEEEauthorblockA{\textit{Fraunhofer IOSB} \\
Karlsruhe, Germany \\
juergen.beyerer@iosb.fraunhofer.de}
}

\maketitle

\begin{abstract}
   Few-shot object detection (FSOD) has thrived in recent years to learn novel object classes with limited data by transferring knowledge gained on abundant base classes. FSOD approaches commonly assume that both the scarcely provided examples of novel classes and test-time data belong to the same domain. However, this assumption does not hold in various industrial and robotics applications, where a model can learn novel classes from a source domain while inferring on classes from a target domain. In this work, we address the task of zero-shot domain adaptation, also known as domain generalization, for FSOD. Specifically, we assume that neither images nor labels of the novel classes in the target domain are available during training. Our approach for solving the domain gap is two-fold. First, we leverage a meta-training paradigm, where we learn the domain shift on the base classes, then transfer the domain knowledge to the novel classes. Second, we propose various data augmentations techniques on the few shots of novel classes to account for all possible domain-specific information. To constraint the network into encoding domain-agnostic class-specific representations only, a contrastive loss is proposed to maximize the mutual information between foreground proposals and class embeddings and reduce the network's bias to the background information from target domain. Our experiments on the T-LESS, PASCAL-VOC, and ExDark datasets show that the proposed approach succeeds in alleviating the domain gap considerably without utilizing labels or images of novel categories from the target domain.
\end{abstract}

\begin{IEEEkeywords}
domain adaptation, few-shot object detection
\end{IEEEkeywords}

\section{Introduction}
Autonomous robotic systems are required to understand and interact with their surrounding environment. With the advent of deep learning, object detection (OD) is a core functional module in the robot's perception pipeline. In various industrial applications, a situation may arise where novel classes are encountered and ought to be learned, such as a pick and place, where a robot is tasked to handle new objects. However, acquiring meaningful samples can be limited, impeding the training of a data-hungry object detector. Moreover, the few provided examples of novel classes might be drawn from a different domain distribution than those encountered at test-time. For instance, the new classes during training can be provided in the form of CAD-rendered images only, which entails a synthetic-to-real domain gap. Another possible scenario would feature data that might have been collected with a camera sensor in a limited environment (for example, by taking mobile images of objects on a table), while the robot should detect the objects for example with a different in-hand camera. Assuming the same domain distribution for training and test-time may result in a suboptimal detection accuracy. 

 \begin{figure}[t!]
 \centering
 \includegraphics[width=0.9\linewidth]{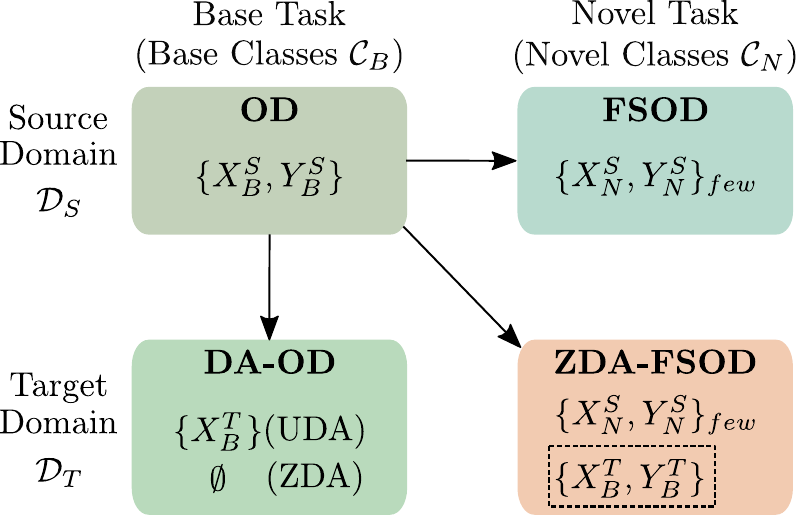}
 \caption{A highlight of the difference among various related learning problems: (i) object detection (OD), (ii) few-shot object detection (FSOD), (iii) domain-adaptive object detection (DA-OD), and (iv) our proposed zero-shot domain adaption task for few-shot object detection (ZDA-FSOD). The available data for each problem during training is shown inside each block. The dashed box denote that the data may not be available.} %
\vspace{-2 em}
 \label{fig:teaser}%
\end{figure}
In this paper, we address the task of detecting novel objects in a target domain, given limited data in a source domain. This complex task is at the intersection of two learning paradigms: few-shot learning (FSL) and domain adaptation (DA) - see Fig.~\ref{fig:teaser}. Note ZDA-FSOD is different from zero-shot object detection (ZOD) \cite{any, zod}, which seeks to detect  unseen classes at test-time in the same source domain. ZDA-FSOD is also different from ZDA, where all classes are seen and available in abundance during training. In contrast to both settings, ZDA-FSOD assumes the presence of few-shots of novel classes in a source domain during training, and seeks to detect these classes in a different domain during testing. FSL~\cite{GFSL} addresses the data scarcity problem by transferring knowledge from base classes with abundant data to novel classes with scarce data, in an attempt to mimic the human cognitive ability to learn new concepts from limited examples. Few-shot object detection (FSOD) is a sub-discipline of  FSL. A typical approach for FSOD is based on meta-learning~\cite{DANA, CME, FsDetView, FSOD-RPN, MetaRCNN, FSRW, MetaDet, DCNET}, which seeks to rapidly learn novel classes by leveraging class-agnostic representations extracted in an episodic training manner. In each episode, different tasks are being solved, where a task may be the localization of different objects in a set of query images, given a small set of support images of the same classes. However, the aforementioned FSOD methods operate under the assumption that train and test images share the same data distribution. This assumption is often violated in the real world,  where multi-level domain shifts on both the image-level (e.g., texture, illumination, resolution and background clutter) and instance-level (e.g., object size and appearance) can occur. Current meta-detectors are not optimized to operate across domains, and are susceptible towards encoding domain-specific and class-specific information as it has been shown that convolutional networks are often biased towards texture \cite{texture_bias}.

Unsupervised domain adaptation (UDA)~\cite{da-survey, uda-survey} is a closely related learning paradigm that transfers knowledge from abundant labeled source domain, $\Ds = \{\mathcal{X}_S, \mathcal{Y}_S\}$, to a label-scarce target domain, $\Dt = \{\mathcal{X}_T\}$. $\mathcal{X}$ and $\mathcal{Y}$ denote the input data samples and the labels, respectively. Zero-shot domain adaptation (ZDA)~\cite{ Kodirov-uda-zero-shot-Domain, zdda-zero-shot-Domain, cocogan-zero-shot-Domain}, also known as domain generalization, further assumes that the data samples in the target domain, $\mathcal{X}_T$ are not available. ZDA thrives to learn domain-agnostic features on multiple source domains \cite{zda_invariant_features}, or to learn the domain shift from an irrelevant task and transfer it to the task of interest using only source domain data \cite{cocogan-zero-shot-Domain, COGANTOI,zdda-zero-shot-Domain}. Nevertheless, these frameworks were designed for the classification task and are not applicable in FSOD for two reasons. First, the existence of paired samples in the irrelevant task is often assumed, which is rarely the case in OD datasets. Second, they involve generative adversarial networks (GANs) \cite{GAN} to synthesize samples of individual objects in the target domain. However, OD datasets contain complex scenes with many objects per image, which in turn imposes a great challenge on GANs to generate meaningful images. Besides DA, domain randomization (DR)~\cite{sim2real-survey} also shares a similar interest, where it mimics the possible physical world scenario through various object simulations (e.g., illumination, pose, size) to learn from a rich data distribution.

Our contributions in this work can be summarized as follows: (1) we introduce the zero-shot domain adaption task for FSOD (ZDA-FSOD),  where  we  learn  to  detect  novel classes in a target domain given only limited samples from the source domain. This problem is different from Zero-Shot Object Detection and Zero-Shot Domain Adaptation for Object Detection. (2) We propose various pixel-level domain randomization techniques (DR) on the query and support data in the novel task. While DR techniques have been previously proposed in \cite{domainrand-2-augmentedae, domainrand-1, domainrand-3, domainrand-4, dtoid}, they are only proposed to address a synthetic-to-real domain gap and use a physics-based blender to augment the data. In contrast, our method leverages augmentations on the present data without generating additional data from a simulator and can be applied on different domain gaps. (3) To promote domain-agnostic class-specific feature embeddings, the mutual information between foreground proposals from the query image and the associated class embedding is maximized by encouraging their embeddings to be closer in the feature space while being further apart from a negative class embedding. (4) Furthermore, we propose to augment class embeddings by sampling from a Gaussian distribution in the feature-level to enrich the data prior to the contrastive loss. (5) We perform experiments on the T-LESS~\cite{tless} and ExDark~\cite{Exdark} dataset to validate the effectiveness of our approach.

 \begin{figure*}[t!]
 \centering
 \includegraphics[width=1.0\linewidth]{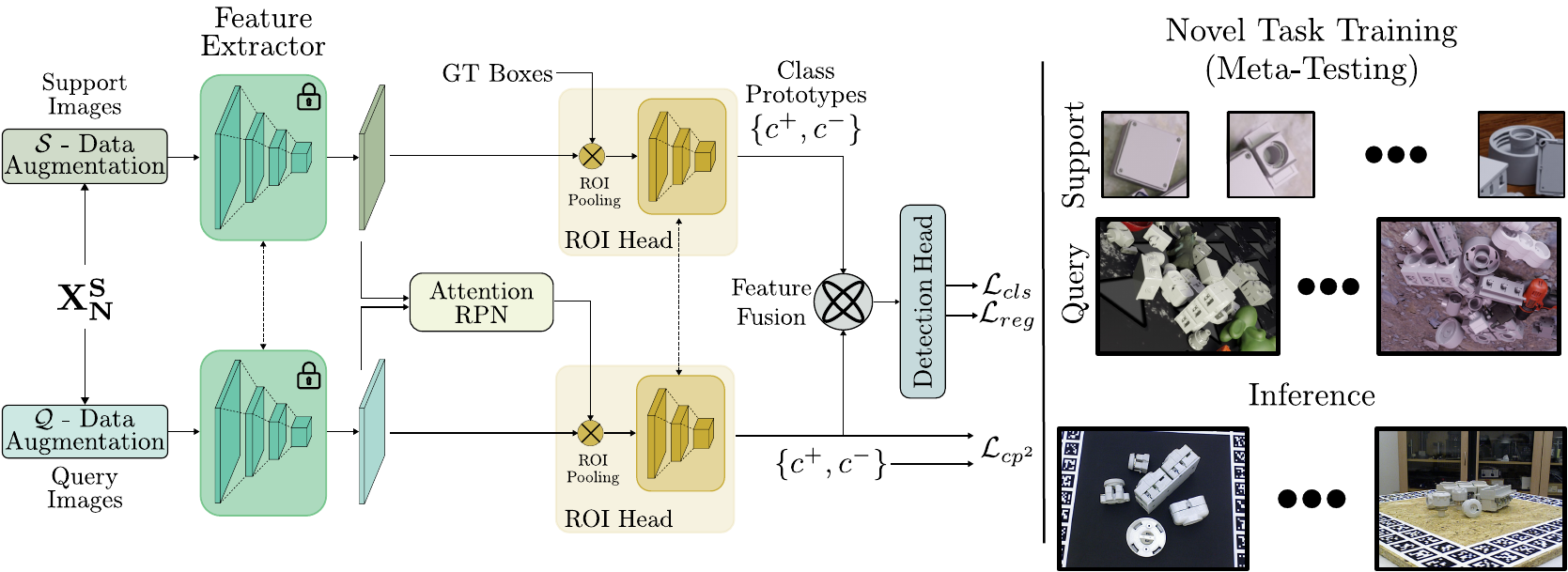}
 \caption{\textit{Left}-~A depiction of the adopted framework for the proposed ZDA-FSOD task, during the novel task training (meta-testing) phase. Given the novel source data $\mathbf{X_N^S}$, the query and support level augmentations are performed and then fed to the network. \textit{Right}-~An illustration of the support and query images during the meta-testing phase (upper) and the inference stage (lower). The Support images used during inference are the ones used during the meta-testing belonging to the source domain.}
 \vspace{-1em}
 \label{fig:architcture}%
\end{figure*} 

\section{Related Works}
\subsection{Few-Shot Object Detection}
FSOD methods are either transfer learning or meta-learning based. Firstly, the transfer learning based few-shot detectors~\cite{LSTD, RepMet, TFA, MPSR, defrcn} aim to transfer knowledge from base to novel classes via fine-tuning. Conversely, meta-detectors strive to extract knowledge across different detection tasks in order to generalize better on new tasks. Meta-detectors can be categorized into: learn to measure~\cite{RepMet, FSRW, MetaRCNN, FSOD-RPN, FsDetView, CME} and learn to finetune~\cite{MetaDet}. The latter performs an instance-level exemplar search via a support set with their features being fused with the query features. Conversely, the former seeks to learn category-agnostic parameters that enable the learning of novel category-specific weights on the new task~\cite{MetaDet}. The models mainly differ in the location of fusion as well as the fusion technique and the training strategy. MetaYOLO~\cite{FSRW} performs feature fusion directly prior to the detection head. As for sparse detectors e.g., MetaRCNN~\cite{MetaRCNN} and FsDetView~\cite{FsDetView}, the feature fusion occurs at the instance-level after the region proposal network (RPN). However, FSOD-RPN~\cite{FSOD-RPN} performs fusion before the RPN to filter irrelevant proposals. Moreover, it is the only framework which employs a two-way contrastive training strategy. CME \cite{CME} maximizes the margin space between base classes to accommodate the novel classes while seeking margin equilibrium during fine-tuning to avoid high variations in novel features. 

\subsection{Domain Adaptation}
Recent domain adaptation methods via label-scarce or none target domain data can be summarized into three main approaches: unsupervised domain adaptation (UDA), domain randomization (DR), and zero-shot domain adaptation (ZDA). UDA~\cite{uda-survey} strives to transfer knowledge across various domains by mitigating the distributional variations. Commonly, UDA methods attempt either a feature-level adaptation in latent space or a pixel-level adaptation via image-to-image translation techniques. Chen \etal proposed domain adaptive Faster R-CNN (DA-FRCNN)~\cite{DAFRCNN} that integrates Faster R-CNN~\cite{FasterR-CNN} with adversarial training to align both the image and instance distributions across domains whilst applying a consistency regularization to learn a domain-invariant RPN. A variety of adversarial-based methods later followed~\cite{A-DAOD1, FAFRCNN, SWDA, A-DAOD2, A-DAOD3, A-DAOD4, A-DAOD5}.

DR~\cite{sim2real-survey}, on the other hand, learns domain-invariant features via rendering images with random illumination, pose, background, etc, which attempts to mimic the real world distribution. Defying the common assumption of task-relevant target domain data availability, ZDA~\cite{Kodirov-uda-zero-shot-Domain,  zdda-zero-shot-Domain, cocogan-zero-shot-Domain} leverages task-irrelevant dual-domain pairs with neither images nor labels from the task-relevant target domain. Recently, various DR methods were introduced in robotics applications like 6D object detection ~\cite{domainrand-2-augmentedae}, 6D object tracking~\cite{domainrand6dtracking}, object localization~\cite{domainrand-1}, person detection~\cite{domainrand-4} and segmentation ~\cite{domainrand-3}. Nevertheless, the aforementioned DR methods mostly rely on a blender or a game engine to generate semi-realistic images. In contrast, we introduce pixel-level and feature-level data augmentations which are applicable to any domain gap.

\section{Methodology}
\label{sec:method}
In this section, we formulate the zero-shot domain adaptation problem for few-shot 2D object detection before presenting our approach. We define the base task as $\mathbf{T}_B = \{ X_B, Y_B, P(Y_B|X_B) \}$ and the novel task as $\mathbf{T}_N = \{ X_N, Y_N, P(Y_N|X_N) \}$, where $X$ refers to the data samples and $Y$ refers to the set of object annotations comprising bounding box coordinates and class labels. The base classes $\mathcal{C}_B$ and novel classes $\mathcal{C}_N$ are disjoint (i.e., $\mathcal{C}_B~\cap~\mathcal{C}_N = \emptyset$). In ZDA, it is often assumed that abundant base target domain data is available and labeled, that is $X_B = \{ X_B^S, X_B^T \}$ and $Y_B = \{ Y_B^S, Y_B^T \}$, where $X^S$ and $X^T$ refer to the source and target domain data samples, respectively. $Y^S$ and $Y^T$ represent the set of object annotations for the source and target domain data samples, respectively . In contrast, in the novel task only a few-shot of source domain data is available, $X_N = \{X_N^S\}$ and $Y_N = \{Y_N^S\}$. In this work, we impose a harsher constraint, and assume that target data is not available in the base task, or is only available with few-shots. The former setting is harder than the latter since no knowledge about the target domain is present at all. 

 \begin{figure*}[t!]
 \centering
 \includegraphics[width=1.0\linewidth]{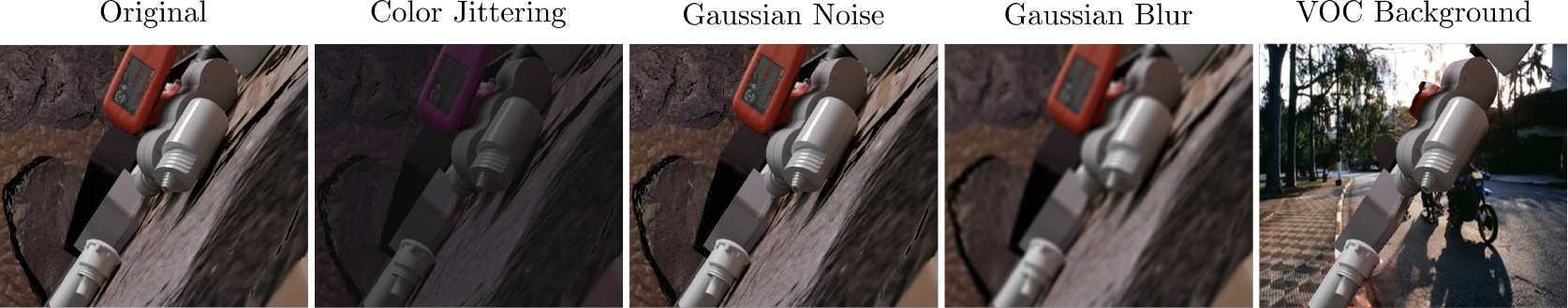}
 \caption{An illustration of various data augmentations applied to a query image from a T-LESS source domain image.}
 \vspace{-1em}
 \label{fig:domain_rand}%
\end{figure*} 
\subsection{Baseline Description}
Our approach extends the FSOD-RPN~\cite{FSOD-RPN} framework to tackle the ZDA-FSOD task. FSOD-RPN \cite{FSOD-RPN} is a two-stage meta-detector built upon Faster-RCNN \cite{FasterR-CNN}. Given a query image, and support images, FSOD-RPN extracts a class embedding from the support images and seeks to detect the objects in the query that belong to the same class. Specifically, it modifies the RPN to filter irrelevant object proposals by fusing the query and support features. The detection head is also modified to integrate a fusion between the query and support feature embeddings via concatentation, followed by an attention mechanism on a global, pixel and patch level. We opt for FSOD-RPN~\cite{FSOD-RPN} as our baseline since it was the state-of-the-art meta-detector at the start of this work.  The baseline along with our contributions in the meta-testing phase is shown in Fig.~\ref{fig:architcture}. 

\subsection{Data Augmentation Schemes }
\label{sec:augmentations}

Since the learning of novel classes is conducted in a supervised manner, the data scarcity makes the model highly prone to overfitting on the domain information. To this end, we propose to leverage various data augmentation techniques during the meta-testing phase. Contrary to previous approaches in domain randomization~\cite{domainrand-1, domainrand-2-augmentedae, domainrand-3, domainrand-4}, we do not use a blender or a game engine to generate multiple poses and lighting conditions. Instead, we leverage pixel-level and feature-level augmentations which allow for the proposed method to tackle any domain gap, rather than being restricted to a sim2real application. The following augmentations are proposed:

\textbf{Color Jittering.} To account for diverse point-wise lighting variations, we randomly apply color jittering by changing the brightness, contrast, hue, and saturation of both the query and support images. This, in turn, helps the network to be less prone to overfitting the source domain colors while learning more domain-invariant features. 

\textbf{Gaussian Blur.} In case of a low-light captured image is noisy or a an out-of-focus frame is encountered, the resulting image can be confusing to the detector. Consequently, we propose to randomly apply Gaussian blur with random kernel sizes and standard deviations on both the support and query images to account for such distortions. 

\textbf{Gaussian Noise.} The scarcity of novel examples can generate a highly noisy learning signal resulting in adverse consequences on the computed features. As a remedy, we randomly add Gaussian noise with various small standard deviation values on both the support and query images. This augmentation was inspired from the work on adversarial attacks on neural networks \cite{adversarial_attack}, where it was found that even a small imperceptible perturbations can fool the network's decision and push a sample away of the classifier's decision boundary. The aim is to encourage the network to be less sensitive to the absolute pixel values of the few available support and query images, in order to decrease the number of false positives and negatives. The increased network's robustness against pixel perturbations results in a higher transferability across domains. 

\textbf{Background Augmentation.} Training on abundant source domain data may result in overfitting the backgrounds of the training images, which are non-transferable across domains. In contrast, foreground object classes should have the same feature representation across domains. To avoid domain confusion caused by a different background at test-time, we propose to extract the objects from the given source query images using their masks or bounding boxes and paste it on real background images using the PASCAL VOC~\cite{pascalvoc} 2007 and 2012 datasets, similar to \cite{domainrand-2-augmentedae}. Rather than applying this augmentation to a single object, we opt to apply it on the query image which contain multiple foreground objects. Specifically, an image is randomly sampled from the VOC dataset and resized to the resolution of the source query image. The extracted objects are then placed at the exact same location in the VOC image to keep the bounding box labels unchanged.  
The aforementioned data augmentations are presented in Fig.~\ref{fig:domain_rand}.

\subsection{Mixed Domain Training Strategy}
\label{sec:mixed}
To avoid overfitting on domain-specific features, we hypothesize that the model should be exposed to more than one domain during meta-training as well. Since we assume that no target domain data is available in the base task, we leverage data augmentations presented in Section \ref{sec:augmentations} on the source data. The training set becomes $\{(X_B^S)^t, Y_B^S\}$, where $t \in \mathcal{T}$ is a transform. The augmentations force the model to observe more than one domain at this stage, since the feature extractor is frozen in the next stage and transfers the learned knowledge on the base task to the novel tasks. 

In case where few-shots of target domain data are present in the base task, we leverage the meta-learning paradigm by redefining the episodic base tasks during the meta-training phase. Rather than solely learning the base tasks on source domain data, instead we augment the base source data with a few-shots from the target domain (i.e. $\{(X_B^S, Y_B^S), (X_B^T, Y_B^T)\}$), to extract cross-domain knowledge. We propose to perform mixed domain training on both the query and support levels. Specifically, during each episode in the meta-training, the network observes query images drawn from either the source or target domain, while the support images might be sampled from a single domain or a mixture of both. The proposed mixed domain training strategy (MDTS) allows the network to observe visual cues in the query image that are less domain-dependent and to transfer the acquired domain knowledge to the novel task by freezing the feature extractor and fine-tuning the region-of-interest (RoI) and detection heads.

\subsection{Contrastive Foreground-Class Embedding Loss}
\label{sec:cfcel}
Learning discriminative features for the novel classes across domains with limited data samples is quite challenging. An often observed drawback is that the model becomes highly confused (i.e., increased false positives/negatives) between classes with similar appearance, or between foreground and background in the new domain. A crucial property of a robust feature representation is that it should encode class-specific domain-agnostic information, meaning it should only be sensitive to the shape of the foreground object. To enforce semantic consistency between foreground and background embeddings, we propose a contrastive foreground-class embedding loss (CFCE) between the foreground proposals from the query image and both the positive and negative support class embeddings. Contrastive losses have been successfully utilized to map two different views of the same scene to a similar point in the representation space~\cite{facenet, koch2015siamese, simclr, sup_cont_loss}. In this work, they are utilized to maximize the mutual information between instance-level features in the query and class-embeddings from the support images, regardless of their augmentations. Our aim is to make the decision boundary of the detector biased towards the topology and semantics of the object classes and not to the domain information. 
 \begin{figure}[t!]
 \centering
 \includegraphics[width=1.0\linewidth]{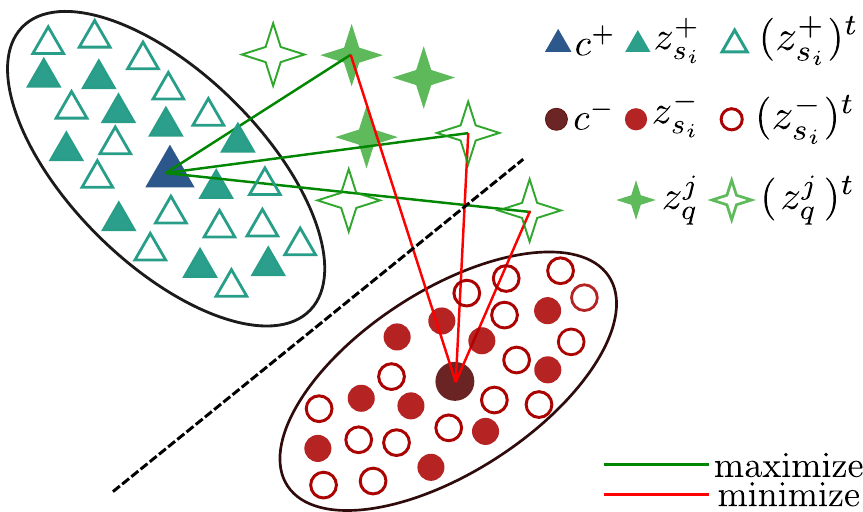}
 \caption{An illustration of the proposed CFCE contrastive loss. The key objective is to maximize the similarity between the foreground query proposals features $z_q^j$ and the positive support class prototype $c^+$, and minimize the similarity between $z_q^j$ and the negative support prototype $c^-$. Moreover, the utilization of augmented features denoted by  $(.)^t$ complements the CFCE loss in order to encourage learning robust cross-domain features.} %
 \label{fig:cfce}%
\end{figure}

We define the features of the foreground query proposals as $\{z_{q}^{j}\}_{j=1}^{P_f}$, where $P_f$ is the number of foreground proposals. Let $\{z_{s_i}^{+}\}_{i=1}^K$ denote positive supports features (that should be detected in the query) while $\{z_{s_i}^{-}\}_{i=1}^K$ denotes negative support features (a randomly sampled class not to be detected). $s_i$ denotes the $i^{th}$ support image and $K$ is the total number of support shots per class. $c^+$ and $c^-$ are the positive and negative class embeddings and are computed by averaging $\{z_{s_i}^{+}\}_{i=1}^K$ and $\{z_{s_i}^{-}\}_{i=1}^K$, respectively, as in our baseline \cite{FSOD-RPN}. Our aim is to map foreground proposals of the query image $q$ within the vicinity of $c^+$ while pushing them away from $c^-$. Specifically, we utilize a triplet margin contrastive loss, which attempts to shorten the distance between $z_{q}^{j}$ and $c^+$ smaller, and to make the distance between $z_{q}^{j}$ and $c^-$ larger than a hyperparameter margin $m$. The triplet margin loss enables a more relaxed feature space that accommodates the inter-class distances~\cite{metriclearning}. We opt for using the cosine similarity distance instead of Euclidean distance, because two similar data objects whose features have a large magnitude in feature space, will have a large Euclidean distance but a smaller angle between them. The CFCE loss is expressed as follows:
\begin{gather}
      \mathcal{L}_{CFCE} = \frac{1}{P_f}\sum_j^{P_f}\max\left(d(z_{q}^j, c^{+}) - d(z_{q}^j, c^{-}) + m, 0\right), \\
      d(v_1, v_2) = \frac{v_1 \cdot v_2}{\parallel v_1\parallel \parallel v_2\parallel}.  
\end{gather}
A visualization of the proposed CFCE loss is presented in Fig.\ref{fig:cfce}.
\subsection{Feature-Level Class Embedding Augmentations}
\label{sec:gaussian}
We observe that computing the class embeddings by directly averaging the $K$-support shot features might not depict the true class embedding distribution due to the scarcity of support shots. Moreover, if the standard deviation between support features is high, the averaging might yield a less informative class embedding. To tackle the aforementioned issues, we propose a data augmentation scheme on the support feature level. Specifically, assuming that the support feature representation follows a Gaussian distribution, we compute the mean feature $\Bar{f}$ over the $K$-shots, and their standard deviation, $\sigma_f$. During each iteration in the meta-testing phase, a latent vector is sampled from the Gaussian distribution $\mathcal{N}(\Bar{f}, \sigma_f^2)$ and is considered as the class embedding. This feature-level augmentation strategy is also implemented during inference to enhance the class embeddings. 

The contrastive losses and data augmentations complement each other. The augmentations introduce perturbations in the feature space enabling the network to explore more examples. A data augmentation might even push an object beyond the detector's decision boundary, simulating the effect of a new domain during inference (see Figure~\ref{fig:cfce}). In our experiments, we observe that strong augmentations can sometimes destabilize the training on the novel task. On the other hand, the contrastive losses attract these far-away features back together by enforcing a semantic consistency, which enables a smoother training and a higher average precision. 

\section{Experimental Setup}
\begin{table}[t!]
	\caption{Summary of the utilized T-Less objects during meta-training (base task) and meta-testing (novel task).\label{tab:dataset}}
	\centering
	\def\arraystretch{1.1}
	\begin{adjustbox}{width=0.9\columnwidth}
		\begin{tabular}{c|c|c}
			\toprule
			Split & Objects & \#Classes\\
			\hline
			Base Task & [1,2,..,18,27] &  19 \\
			Novel Task & [19,20,..,26,28,29,30] &  11\\
			\bottomrule
		\end{tabular}
	\end{adjustbox}
	\vspace{-2em}
\end{table}
\subsection{Dataset}
To evaluate the proposed ZDA-FSOD, a dataset featuring multiple domains is required. Due to the nature of this problem, the most suitable publicly available datasets are T-LESS~\cite{tless}, PASCAL-VOC~\cite{pascalvoc}, and ExDark~\cite{Exdark}.

\textbf{T-LESS Synthetic $\rightarrow$ Real}. T-LESS comprises $30$ industrial objects without any notable texture,  distinct color, or reflectance features. There are two available sets of data: synthetic rendered images including a total of $50$ scenes with $\sim 1000$ images each (source domain) and real images featuring $20$ different scenes (target domain). For the real data, we use only the data captured by the Primesense CARMINE $1.09$ RGB-D sensor. However, all the experiments utilize only the RGB data without incorporating any depth information. To prepare the T-LESS dataset to the FSOD task, we propose a train and test splits for the base and novel classes. We split the T-LESS objects into $19$ base classes and $11$ novel classes. A detailed summary of the utilized splits is presented in Table \ref{tab:dataset}. Moreover, we split the 20 scenes of real images into 8 training scenes, $(2,3,5,6,7,9,11,12)$, featuring mainly the base classes, and the remaining $12$ containing novel training and testing scenes. The inference is done on unseen classes featuring the $11$ novel objects. We report the inference results with $K=5$ and $K=10$.

\textbf{PASCAL-VOC (day) $\rightarrow$ ExDark (night)}.PASCAL-VOC dataset comprises 20 categories. We split into 15 base classes and 5 novel classes. The novel classes are boat, bottle, bus, cat, and motorbike. The data is sampled from the VOC 2007 and VOC 2012 train/val set for base and novel training. As for the ExDark, it comprises 12 classes with 10 overlapping with the PASCAL-VOC dataset. We split ExDark into 7 and 5, base and novel classes, respectively. The novel classes are the same as PASCAL-VOC, while the rest are considered as base classes. The ExDark test set with only the novel classes is utilized for testing. The results are reported for $K=5$ and $K=10$ shots.

\subsection{Evaluation Metrics}
The utilized evaluation metrics are average precision (AP) and average recall (AR) averaged over $10$ intersection over union (IoU) thresholds ranging from $0.5$ to $0.95$ with a $0.05$ step. The reported AP and AR values are averaged across all classes. Moreover, we report $AP50$ and $AP75$ denoting the AP at $IoU=0.5$ and $IoU=0.75$, respectively. 

\subsection{Implementation Details}
For a fair quantitative comparison, we meta-train all models on the base task for $5$ epochs with an SGD optimizer using the default parameters as \cite{FSOD-RPN} and a batch size of $8$. The base learning rate is $0.004$ for the first $3$ epochs and $0.0004$ for another $2$ epochs. For meta-testing, we perform $6$k iterations with a learning rate of $0.001$. Similar to~\cite{FSOD-RPN}, the shorter side of the query image is resized to $600$ pixels while the longer side is resized to $1000$. Moreover, each support image is cropped around the target object with $16$-pixel image context, zero-padded, and then resized to $320 \times 320$.

\section{Results}
\label{sec:results}
\begin{table*}[t!]
\caption{Quantitative Results for T-LESS Synthetic $\rightarrow$ Real. MDTS denotes the mixed domain training strategy (source + few-shot target data) in Section \ref{sec:mixed}. MDTS-Aug denotes source data augmented with DR.}
\vspace{-1em}
\label{table:main_results}
\setlength\tabcolsep{4.0pt}
\begin{center}
\scalebox{1.4}{
\begin{tabular}{c|c|c|cccc|cccc}
	\toprule
  Methods & Meta-Training & Meta-Testing & \multicolumn{4}{c}{5-Shot}  & \multicolumn{4}{c}{10-Shot}  \\
   & Data Domain & Data Domain &AP & AP50 & AP75 & AR & AP & AP50 & AP75 & AR\\
  \toprule
\multirow{4}{*}{Baselines} & (a) Source & \multirow{4}{*}{Source} & 1.7 & 3.0 & 1.8 & 4.1  & 6.1 & 10.1 & 6.5 & 13.3\\
 & (b) Target &  & 6.8 & 12.6 & 6.9 & 19.9  & 10.2 & 16.0 & 11.6 & 21.9 \\
 & (c) MDTS & & 12.9 & 22.1 & 13.0 & 35.2 & 23.5 & 34.0 & 25.7 & 49.0  \\
 \midrule
\multirow{3}{*}{Ours} &  Source  & \multirow{4}{*}{Source} & 6.5 & 11.0 & 6.7 & 32.8 & 17.3 & 27.2 & 18.9 & 44.2 \\ 
 & MDTS-Aug &  & 10.1 & 17.7 & 10.3 & 21.5 & 26.3 & 39.8 & 29.7 & 49.2 \\ 
 & MDTS &  & \textbf{17.4} & \textbf{26.5} & \textbf{18.9} & \textbf{39.9} & \textbf{31.2} &\textbf{45.8} & \textbf{34.0} & \textbf{61.0} \\ \midrule
\cellcolor[HTML]{EFEFEF} Oracle & \cellcolor[HTML]{EFEFEF}Target  & \cellcolor[HTML]{EFEFEF}Target & \cellcolor[HTML]{EFEFEF}37.5 & \cellcolor[HTML]{EFEFEF}50.8 & \cellcolor[HTML]{EFEFEF}42.5 & \cellcolor[HTML]{EFEFEF}53.9 & \cellcolor[HTML]{EFEFEF}51.7 & \cellcolor[HTML]{EFEFEF}68.4 & \cellcolor[HTML]{EFEFEF}58.7 & \cellcolor[HTML]{EFEFEF}62.7 \\
\bottomrule
\end{tabular}
}
\end{center}
\vspace{-1em}
\end{table*}
\begin{table*}[t!]
\caption{Quantitative Results for PASCAL VOC0712 $\rightarrow$ ExDark. MDTS denotes the mixed domain training strategy (source + few-shot target data) in Section \ref{sec:mixed}. MDTS-Aug denotes source data augmented with DR. However, no VOC Background augmentation is used since PASCAL VOC is the source dataset.}
\vspace{-1em}
\label{table:voc_main_results}
\setlength\tabcolsep{4.0pt}
\begin{center}
\scalebox{1.4}{
\begin{tabular}{c|c|c|cccc|cccc}
	\toprule
  Methods & Meta-Training & Meta-Testing & \multicolumn{4}{c}{5-Shot}  & \multicolumn{4}{c}{10-Shot}  \\
   & Data Domain & Data Domain &AP & AP50 & AP75 & AR & AP & AP50 & AP75 & AR\\
  \toprule
\multirow{3}{*}{Baselines} & (a) Source & \multirow{4}{*}{Source} & 10.1 & 26.4 & 5.4 & 16.9 & 12.1 & 32.4 & 6.0 & 20.0\\
& (b) Target & & 5.3 & 16.1 & 1.8 & 11.1 & 6.9 & 21.9 & 2.9 & 14.1\\
& (c) MDTS & & 11.8 & 29.9 & 6.7 & 21.3 & 13.5 & 34.3 & 7.7 & 21.8\\
 \midrule
\multirow{3}{*}{Ours} &  Source  & \multirow{3}{*}{Source} & 11.0 &  28.6 & 6.5 & 20.7 & 13.6 & 34.7 & 9.4 & 23.6\\
 & MDTS-Aug & & \textbf{13.0}& \textbf{32.6} & \textbf{7.4} & \textbf{23.5} & \textbf{14.2} & \textbf{36.7} & \textbf{9.6} & \textbf{25.0}\\
 & MDTS & & 11.9 & 30.6 & 7.1 & 21.5 & 13.9 & 35.5 & 9.0 &  23.4\\ \midrule
\cellcolor[HTML]{EFEFEF} Oracle & \cellcolor[HTML]{EFEFEF}Target  & \cellcolor[HTML]{EFEFEF}Target & \cellcolor[HTML]{EFEFEF} 10.0 & \cellcolor[HTML]{EFEFEF} 24.7 & \cellcolor[HTML]{EFEFEF} 7.2 & \cellcolor[HTML]{EFEFEF} 21.2 & \cellcolor[HTML]{EFEFEF} 14.2 & \cellcolor[HTML]{EFEFEF} 35.4 & \cellcolor[HTML]{EFEFEF} 9.4 & \cellcolor[HTML]{EFEFEF} 25.5\\

\bottomrule
\end{tabular}
}
\vspace{-1em}
\end{center}
\end{table*}

In the following section, we evaluate the proposed approach on  T-LESS~\cite{tless} and ExDark~\cite{Exdark} datasets. 

\textbf{Baselines.} To test the performance of the proposed framework, we consider the four following baselines: (a) FSOD-RPN meta-trained on source data only, (b) FSOD-RPN meta-trained on target data only, and finally, (c) FSOD-RPN meta-trained on source data and few-shot target data. All models are finetuned on novel classes from the source domain and tested on novel classes from the target domain. Note that baseline (b) assumes abundant target data in the base task and thus have an advantage over the proposed model. Note that we do not compare with other FSOD frameworks in this task because they have different model capacities and report different performances on the same datasets, which would present an unfair comparison in a domain adaptation setting. Although a few works~\cite{any, FSRW} have reported cross-datasets results, namely from MS-COCO to PASCAL-VOC, these methods reported the results as an experiment to test generalization on more novel classes, not as a straightforward solution to ZDA. Moreover, there is no apparent domain gap between MS-COCO and PASCAL VOC. 

\begin{table*}[t!]
\caption{Meta-testing on T-LESS dataset with meta-trained weights on mixed domain samples in the base task with our MDTS approach.}
\label{table:meta-test}
\begin{center}
\scalebox{1.4}{
\begin{tabular}{lll|cccc}
	\toprule
  &\multicolumn{2}{c|}{Data Configuration} & \multicolumn{4}{c}{10-Shot Inference}\\
  & \multicolumn{1}{c}{Support} & \multicolumn{1}{c|}{Query} & AP & AP50 & AP75 & AR\\
  \midrule
\textbf{A} &  Source Only & Source Only & 23.5 & 34.0 & 25.7 & 49.0  \\  
\textbf{B} &  + Color Jittering &  + Color Jittering & 26.3 & 38.3 & 27.9 & 59.0  \\  
\textbf{C} & + Gaussian Blur & + Gaussian Blur & 26.8 & 39.9 & 30.0 & 58.1  \\  
\textbf{D} & + Gaussian Noise & + Gaussian Noise & 29.3 & 43.8 & 31.9 & 59.2  \\  
\textbf{E} & & + VOC Background & 30.2 & 44.3 & 32.6 & 57.6  \\ 
\textbf{F} &  + Feature Noise & & 30.4 & 44.8 & 33.4 & \textbf{63.1}  \\ 
\rowcolor[HTML]{EFEFEF}
\textbf{G} & \multicolumn{2}{c|}{+ CFCE Loss}& \textbf{31.2} & \textbf{45.8} & \textbf{34.0} &61.0\\  
\bottomrule
\end{tabular}}
\end{center}
\vspace{-2em}
\end{table*}

\subsection{Main Results}
We test $3$ variants of our model. The first one is meta-trained on source only data, the second one is trained on source data that is augmented with the proposed DR techniques, and the final variant uses the MDTS (source + few-shot target data). The $3$ variants are meta-tested with the proposed DR techniques, the CFCE loss and the feature-level augmentations. As oracle (upper bound), we consider a FSOD-RPN model meta-trained and meta-tested on actual target data.

In Table~\ref{table:main_results}, we report the results on the T-LESS dataset under $K=5, 10$-shot settings. First, we notice that without any DR, the source-only baseline (baseline (a)) exhibits a large performance drop in AP and AR, from $51.7$ to $6.1$, and $62.7$ to $13.3$, respectively. Second, the MDTS in (c) is shown to surpass meta-training on either domain alone (baseline b and c). Specifically, MDTS surpasses even target-only meta-training. This can be attributed to the fact that the model reaches a better optimum when two different data distributions of the same classes are observed during meta-training. 

Third, the proposed DR approach can considerably improve the AP and AR in different meta-training settings. When performed on a source-only meta-trained model, DR boosts the AP by $11.2$ points under the $10$-shot setting (row 5). The model can reach an AP of $26.3$, almost $51\%$ of the oracle's performance, without seeing any target domain sample in the meta-training and meta-testing phases (row $5$ in Table~\ref{table:main_results}). This variant already surpasses the MDTS training in baseline (c), while seeing less data. The best performance on $5$ and $10$-shot, $17.4$ and $31.2$, were reached using mixed domain samples in meta-training and DR in meta-testing (row $6$ in Table~\ref{table:main_results}). However, the relative performance gain is higher under the $10$-shot settings ($60\%$ of the oracle) than $5$-shot settings ($46.4\%$ of the oracle) due to the very sparse nature of the latter setting.

It is important to note that we also trained a baseline where both datasets are fully available, but the training did not converge. We hypothesize that either the partially frozen ImageNet pretrained feature-extractor or the fusion mechanism in the Attention-RPN~\cite{FSOD-RPN} is a limiting factor. However, more investigations are needed.  

In Table~\ref{table:voc_main_results}, we report the results on the PASCAL VOC/ExDark datasets, to study the generalization of our model on a different domain gap (day to night). In contrast to T-LESS, we find that meta-training on target (b) does not yield any improvement over source-only training (a). The drop in baseline (b) can be attributed to 2 factors. First, ExDark has less base classes than PASCAL VOC in meta-training. Second, the domain gap is considerably harder to learn in a few-shot setting than that of real-to-synthetic domain gap in T-LESS, because some objects are hardly visible in ExDark and hence are sometimes confused with the background by the network. However, we find that our models can outperform all baselines consistently on this new domain gap, and that the best performance was attained in row 6 (MDTS-Aug), without using any target domain data during meta-training. Note that many models in the table are better than the oracle. We attribute this to the fact that the target data (ExDark) contains less base classes than the source data (PASCAL VOC).

\begin{table*}[t!]
	\caption{A study of cross-domain training settings during meta-training and meta-testing phases on T-LESS dataset. All experiments are conducted using the FSOD-RPN as a baseline. $\checkmark^{few}$ denotes few-shot samples of the base task in the target domain. The gray highlighted rows denote the utilization of our proposed DR approach.}
	\label{meta-study}
    \centering
	\begin{adjustbox}{width=1.0\textwidth}
    	\begin{tabular}{cc|cc|c|cccc|cccc}
        \toprule
         \multicolumn{4}{c|}{Meta-Training} & \multirow{3}{*}{Meta-Testing} & \multicolumn{8}{c}{Inference}\\ 
        \multicolumn{2}{c}{~Support~~} & \multicolumn{2}{c|}{~Query~~} & & \multicolumn{4}{c}{5-shots} & \multicolumn{4}{c}{10-shots} \\ 
        Source & Target & Source & Target & & AP & AP50 & AP75 & AR  & AP & AP50 & AP75 & AR\\ \toprule
    		          
    		          \checkmark   & & \checkmark & & \multirow{4}{*}{Source Only} & 1.7 & 3.0 & 1.8 & 4.1  & 6.1 & 10.1 & 6.5 & 13.3\\
    		          & \checkmark & & \checkmark &  & 6.8 & 12.6 & 6.9 & 19.9  & 10.2 & 16.0 & 11.6 & 21.9\\
    	  	          \checkmark   &  $\checkmark^{few}$ & \checkmark & $\checkmark^{few}$ &  & 12.9 & 22.1 & 13.0 & 35.2 & 23.5 & 34.0 & 25.7 & 49.0\\
    	  	          \rowcolor[HTML]{EFEFEF}
    	  	          \checkmark   &  $\checkmark^{few}$ & \checkmark & $\checkmark^{few}$ &  & \textbf{17.4} & \textbf{26.5} & \textbf{18.9} & \textbf{39.9} & \textbf{31.2} & \textbf{45.8} & \textbf{34.0 }& \textbf{61.0}\\
    	  	          \midrule
    	  	         \checkmark &  & \checkmark & & \multirow{4}{*}{Target Only} & 38.3 & 54.7 & 43.0 & 56.5 & 48.7 & 65.2 & 55.0 & 64.6\\
                	& \checkmark & & \checkmark &  & 37.5 & 50.8 & 42.5 & 53.9 & 51.7 & 68.4 & 58.7 & 62.7\\
    	  	         \checkmark   & $\checkmark^{few}$ & \checkmark & $\checkmark^{few}$ & & 39.9 & 54.0 & 44.6 & 58.9 & 48.3 & 62.4 & 53.7 & 66.1\\
    	  	         \rowcolor[HTML]{EFEFEF}
    	  	          \checkmark   &  $\checkmark^{few}$ & \checkmark & $\checkmark^{few}$ &  & \textbf{39.9} & \textbf{55.2} & \textbf{44.6} & \textbf{62.9} & \textbf{49.2} & \textbf{65.0} & \textbf{54.6} & \textbf{69.7}\\
    		\bottomrule
    	\end{tabular}
	\end{adjustbox}
\end{table*}

\subsection{Ablation Study on Meta-Testing via DR}
In Table~\ref{table:meta-test}, we conduct different experiments to study the impact of different proposed augmentations during meta-testing. In all experiments, we use a meta-trained network on source and few-shot target domain data. Starting from source only meta-testing (configuration \textbf{A}), we incrementally apply augmentations and report the results. First, we note that the color jittering, in configuration \textbf{B}, significantly boosts the AR by $10$ points denoting a steep decrease in the false negatives, and a $2.2$ points improvement in the AP. While introducing the Gaussian blur (\textbf{C}) further improves the AP by $0.6$ points and slightly deteriorates the AR by $0.9$ points, the Gaussian noise (\textbf{D}) enhances the AP by $2.5$ points and recovers the AR by a surplus with a total of $1.1$ points. This illustrates the important role played by introducing the pixel-level perturbations in yielding a more robust model towards simple distortions. Next, in configuration \textbf{E}, we add the VOC background augmentation on the query images, and notice an improved AP. The feature-level augmentations in configuration \textbf{F} slightly increase the AP, although their main effect is reflected in the high AR. Finally, the best result in our experiments was achieved by adding the CFCE loss which boosts the AP by $0.8$ point, but slighlty decrease the AR to $61$, which is close to the oracle's AR, $62.7$. We have also observed that the introduced contrastive loss and feature-level augmentation stabilize the meta-testing, besides decreasing the number of false negatives and positives. 

\section{Ablation Study on Cross-Domain Training}

This set of experiments is designed to explore different cross-domain settings with the aim to analyze the existing domain gaps in meta-training and meta-testing, as a guide to future research in this direction. The proposed DR approach is only used in the highlighted rows in Table~\ref{meta-study}. We can draw 3 conclusions from the results. First, the existence of target-domain samples for novel classes can almost bridge the domain gap, even if no target domain samples were seen during meta-training. Second, the model can still benefit with the MDTS in the meta-training, regardless of the seen samples during meta-testing. This is shown in rows $3, 4, 7, 8$ in Table \ref{meta-study}, where the model can achieve an AR higher than the oracle under the $5,10$-shot settings, a higher AP under the $5$-shot setting and an AP close to the oracle under the $10$-shot setting. Third, even when the DR approach is applied during target-only meta-testing, it improves upon the oracle in $5$-shot settings (AR and AP) and $10$-shot settings (AR). 

However, we note that the existence of target domain samples during meta-testing is an assumption that does not hold in many applications, as there exists often a difference in the background, the camera sensor, or a sim2real domain gap. Additionally, it is not practical to finetune the pretrained model on a new domain, every time a new camera is installed, or a change in the environment occurs.

\subsection{Discussion}
One limitation of the proposed framework is that the CFCE loss is sensitive to hyperparameters. Future works could consider stabilizing the CFCE loss. Overall, the proposed ZDA-FSOD framework has shown a solid generalization to multiple domain gaps (syn-to-real, day-to-night). We believe our contributions can be applied to other FSOD frameworks, however, this is considered future works and is out of the scope of this paper. The FSOD-RPN~\cite{FSOD-RPN} was chosen as it represented the state-of-the-art in FSOD at the start of this work. 

\section{Conclusion}
In this paper, we introduced the task of learning to detect new classes in an unseen domain, given a few-shots from a source domain. This paves the way towards a wide variety of applications in different robotics and industrial use-cases. To this end, we proposed to mimic the domain shifts that may be encountered during the test-time by designing a domain randomization strategy in a meta-learning paradigm. We proposed a contrastive loss along with feature-level augmentations which, besides stabilizing the novel task learning, foster the network to encode domain-agnostic class-specific feature embeddings that are less sensitive to the surrounding environment variations. We also highlighted the effect of meta-training with mixed domain samples, which either include source and target samples or augmented source samples to simulate various domains. The experiments have shown that we successfully readjusted a meta-detector trained using source domain samples, to tackle the existing domain gap for an unseen domain under a few-shot setting. We believe that the proposed method is beneficial to numerous practical real-world robotic autonomous systems and encourages further research on tackling domain shifts when learning novel objects in unseen domains.     

\bibliographystyle{ieee_fullname}
\bibliography{IEEEabrv, egbib}

\end{document}